\documentclass[conference]{IEEEtran}
\IEEEoverridecommandlockouts

\usepackage{cite}
\usepackage{amsmath,amssymb,amsfonts}
\usepackage{algorithm}
\usepackage{algpseudocode}
\usepackage{graphicx}
\usepackage{textcomp}
\usepackage{algpseudocode}

\usepackage[table]{xcolor}
\usepackage{hyperref}

\newtheorem{thm}{Theorem}

\usepackage{multirow}
\usepackage{caption}
\def\BibTeX{{\rm B\kern-.05em{\sc i\kern-.025em b}\kern-.08em
    T\kern-.1667em\lower.7ex\hbox{E}\kern-.125emX}}
    
\usepackage{tabularx}
\usepackage{booktabs}
\usepackage{authblk}
\begin{document}
\title{Parallel feature selection based on the trace ratio criterion}
\author[1]{Thu Nguyen*\thanks{* denotes equal contribution}}
\author[3,4,5]{Thanh Nhan Phan*}
\author[2]{Nhuong Nguyen}
\author[3,4,5]{Thanh Binh Nguyen}
\author[1]{Pål Halvorsen}
\author[1]{Michael Riegler}
\affil[1]{SimulaMet, Norway}
\affil[2]{University of Connecticut, USA}
\affil[3]{AISIA Research Lab, Vietnam}
\affil[4]{University of Science, Vietnam}
\affil[5]{Vietnam National University, Vietnam}
\maketitle
\begin{abstract}
The growth of data today poses a challenge in management and inference. While feature extraction methods are capable of reducing the size of the data for inference, they do not help in minimizing the cost of data storage. On the other hand, feature selection helps to remove the redundant features and therefore is helpful not only in inference but also in reducing management costs. This work presents a novel parallel feature selection approach for classification, namely Parallel Feature Selection using Trace criterion (PFST), which scales up to very large datasets. Our method uses trace criterion, a measure of class separability used in Fisher's Discriminant Analysis, to evaluate feature usefulness. We analyzed the criterion's desirable properties theoretically. Based on the criterion, PFST rapidly finds important features out of a set of features for big datasets by first making a forward selection with early removal of seemingly redundant features parallelly. After the most important features are included in the model, we check back their contribution for possible interaction that may improve the fit. Lastly, we make a backward selection to check back possible redundant added by the forward steps. We evaluate our methods via various experiments using Linear Discriminant Analysis as the classifier on selected features. The experiments show that our method can produce a small set of features in a fraction of the amount of time by the other methods under comparison. In addition, the classifier trained on the features selected by PFST not only achieves better accuracy than the ones chosen by other approaches, but can also achieve better accuracy than the classification on all available features.  
\end{abstract}
\section{Introduction}

In this era of big data, the growth of data poses challenges for effective data management and inference. For example, a gene dataset can contain hundred thousands of features \cite{guyon2007competitive}. Hence, directly handling such datasets may face the curse of dimensionality. Moreover, the presence of redundant features can derail the learning performance of classification algorithms. For dealing with this issue, many dimension reduction techniques have been developed \cite{melab2006grid,de2006parallelizing,garcia2006parallel,guillen2009efficient,lopez2006solving}, and they are categorized into either feature extraction or feature selection methods \cite{liu2012feature,kumar2014feature}.

Feature extraction techniques (e.g., Principal Component Analysis \cite{johnson2002applied}, Linear Discriminant Analysis \cite{johnson2002applied}) involve projecting the data into a new feature space with lower dimensionality via some linear or nonlinear transformation of the original features. However, this creates sets of new features that can not be directly interpreted. Moreover, since those approaches use all the features available during feature extraction, it does not help reduce the cost of data storage and the cost of collecting data in the future. 

On the other hand, feature selection methods (forward selection \cite{james2013introduction}, backward selection \cite{james2013introduction}, etc.) select only a subset of useful features for model construction. Therefore, it maintains the original features' meanings while reducing the cost of storage and collecting data in the future by removing irrelevant features. 


However, data from various fields, such as text mining, business analytics, and biology, are often measured in gigabytes or terabytes with millions of features \cite{bolon2015feature, li2017feature}. For example, the Amazon Review dataset \cite{ni2019justifying} is a 34 gigabytes dataset. 
In such cases, the performance of the most current feature selection techniques may be jeopardized \cite{li2017feature}. This is due to the search space for a set of relevant features increases significantly. One way to deal with this issue is to go for parallelization, which allows better use of computers' computational resources by data partitioning and running features selection on multiple cores at the same time. 

In this work, we develop a novel parallel feature selection method for classification that can help remove redundant features and improve the model's performance without sacrificing the running time for feature selection. This is achieved by using the trace criterion in Linear Discriminant Analysis \cite{johnson2002applied} as the evaluation criteria for the measure of separability between classes and therefore help evaluate if adding a feature helps to improve the model.    
Using this criterion, we propose a parallel feature selection approach, namely Parallel Feature Selection based on Trace criterion (PFST), which consists of three stages. First, in a forward-dropping stage, we do parallel forward selection on multiple workers and discard seemingly redundant features from subsequent steps of the stages. At the second stage, after the most important features are already included in the model, we check back their contribution after including many essential features for possible interaction that may improve the fit. Finally, we conduct backward selection in the last stage to check back possible irrelevant features added by the forward moves. 

The rest of the article is organized as follows: First, Section \ref{related} gives a review of related works in feature selection. Second, Section \ref{prelim} reviews some basic approaches in feature selection, and Section \ref{tracecri} details the trace criterion and some of its desirable properties as a measure of class separability for feature selection. Next, Section \ref{methods} describes our methodology, and Section \ref{exper} illustrates the power of our method via experiments on various datasets. Lastly, we summarize our approach and discuss future works in Section \ref{conclu}. 

\section{Related Works} \label{related}

Due to the benefits of maintaining original features while reducing the storage cost of feature selection, there have been many efforts in the field of feature selection to develop new techniques for the growing data size challenges. Aside from hybrid approaches that combine different feature selection strategies \cite{saeys2007review,ang2015supervised}, most feature selection methods can be classified into three categories.  

First, the wrapper approaches rely on the performance of a specified learning algorithm to evaluate the importance of selected features. A typical wrapper method will first search for a subset of features for a given learning algorithm and then evaluate them. These steps are repeated until some stopping criteria are satisfied. Methods in this category are usually computational expensive as subset evaluation requires multiple iterations. Even though many search strategies, such as best-first search \cite{arai2016unsupervised} and the genetic algorithm \cite{goldberg1988genetic}, have been proposed, using these strategies for high dimensional data is still computational inefficient.

On the other hand, the filter approaches consist of techniques that evaluate feature subsets via ranking with some criteria such as information criteria \cite{nguyen2014effective, shishkin2016efficient}, reconstruction ability \cite{farahat2011efficient, masaeli2010convex}. These approaches choose features independently from learning algorithms and are typically more computationally efficient than wrapper methods \cite{li2017feature}. However, since the are not optimized for any target learning algorithm, they may not be optimal for a specific learning algorithm. 

Meanwhile, the embedded approaches use independent criteria to find optimal subsets for a known cardinality. After that, a learning algorithm is used to select the final optimal subset among the optimal ones across different cardinalities. Therefore, they are more computationally efficient than wrapper methods since they do not evaluate feature subsets iteratively. In addition, they are also guided by the learning algorithm. Therefore, they can be seen as a trade-off between the filter strategy and wrapper strategy \cite{li2017feature}.

Despite efforts in the field of feature selection so far, data from various fields may be too abundant even for filter methods to be computationally efficient. This motivates many types of research in parallel feature selection. For example, the methods proposed in \cite{melab2006grid,de2006parallelizing,garcia2006parallel,guillen2009efficient,lopez2006solving} use parallel processing to evaluate multiple features or feature subsets simultaneously, and therefore speed up the feature selection process. Yet, these algorithms require access to the whole data. 
On the other hand, in \cite{singh2009parallel}, the authors propose a parallel feature selection algorithm for logistic regression based on the MapReduce framework, and features are evaluated via the objective function of the logistic regression model. Meanwhile, the authors in \cite{tsamardinos2019greedy} propose a \textit{Parallel, Forward–Backward with Pruning algorithm} (PFBP) for feature selection by Early Dropping \cite{tsamardinos2019greedy} some features from consideration in subsequent iterations, Early Stopping \cite{tsamardinos2019greedy} of consideration of features within the same iteration, and Early Return of the winner in each iteration. However, this approach requires bootstrap computations of p-value, which is computationally expensive. 
Zhao et al.\cite{zhao2013massively} introduce a parallel feature selection algorithm that selects features based on their abilities to explain the variance of the data. Yet, in their approach, determining the number of features in the model is based on transforming the categorical labels to numerical values and using the sum of squared errors. Getting the sum of squared errors requires fitting the model and, therefore, the algorithm is still computationally expensive.



\section{Preliminaries}\label{prelim}
Various feature selection algorithms have been developed up to now. However, forward, backward, and stepwise feature selection remains widely used. This section summarizes these three techniques, and the details are given in Algorithms \ref{alg:ffs}, \ref{alg:bfs}, and \ref{alg:sfs}.

Forward feature selection (Algorithm \ref{alg:ffs}) has been used widely due to its computational efficiency, along with the ability to deal with problems where the number of features highly exceeds the number of observations efficiently. However, some features included by forwarding steps may appear redundant after including some other features. About the sufficient conditions for forward feature selection to recover the original model and its stability, we refer to \cite{tropp2004greed} and \cite{donoho2005stable} for further readings. 

\begin{algorithm}
\caption{\textbf{Forward Feature Selection}}\label{alg:ffs}
\hspace*{\algorithmicindent} \textbf{Input:} A dataset of $p$ features $f_1,f_2,...,f_p$, threshold $\alpha$.\\
    \hspace*{\algorithmicindent} \textbf{Output:} A set R of relevant features.
\begin{algorithmic}[1]
\State $R \gets \emptyset$
\State $S \gets \{f_1, f_2,\ldots, f_p\}$
\While {True}
    \State $f_j \gets$ the most useful feature in $S$
    \If{the model improves more than an amount of $\alpha$ after including $f_j$}
    \State $R \gets R \cup \{f_j\}$
    \State $S \gets S\setminus \{f_j\}$
    \Else
    \State \Return R
    \EndIf
\EndWhile
\end{algorithmic}
\end{algorithm}

	
	








Backward feature selection \cite{james2013introduction}, as shown in Algorithm \ref{alg:bfs}, starts with the full model containing all the predictors and then iteratively removes the least useful one, one-at-a-time. It ensures that only redundant features are removed from the model. However, backward feature selection is slow compared to forwarding selection. 

\begin{algorithm}
\caption{\textbf{Backward Feature Selection} }\label{alg:bfs}
\hspace*{\algorithmicindent} \textbf{Input:} A dataset of $p$ features $f_1,f_2,...,f_p$, threshold $\beta$.\\
    \hspace*{\algorithmicindent} \textbf{Output:} A set R of relevant features.
\begin{algorithmic}[1]
\State $R \gets \{f_1, f_2,\ldots, f_p\}$
\While {True,}
    \State $f_j \gets$ the least useful feature in R
    \If {the amount of loss by excluding $f_j$ is less than $\beta$}
    \State $R \gets R\setminus \{f_j\}$
    \Else 
        \State \Return R
    \EndIf
    \EndWhile
\end{algorithmic}
\end{algorithm}

	
	








As another alternative, a hybrid version of both forward and backward feature selection is stepwise selection, detailed in Algorithm \ref{alg:sfs}. In this approach, the features are added to the model sequentially as in the forwarding feature selection. However, after adding each new feature, the method may remove any feature that no longer seems applicable.  

\begin{algorithm}
\caption{\textbf{Stepwise Feature Selection} }\label{alg:sfs}
\hspace*{\algorithmicindent} \textbf{Input:} A dataset of $p$ features $f_1,f_2,...,f_p$, forward threshold $\alpha$, backward threshold $\beta$.\\
    \hspace*{\algorithmicindent} \textbf{Output:} A set R of relevant features.
\begin{algorithmic}[1]
\State $R \gets \emptyset$
\State $S \gets \{f_1, f_2,\ldots, f_p\}$
\While {True,}
   \State $f_j \gets$ the most useful feature in $S$
   \If{the model improves more than an amount of $\alpha$ after including $f_j$}
    \State $R \gets R \cup \{f_j\}$
    \State $S \gets S\setminus \{f_j\}$
    \While{True}
        \State {$f_k\gets$  the least useful feature in R}
        \If {the model performance decreases with an amount less than $\beta$ after excluding $f_k$}
        \State $R \gets R\setminus \{f_k\}$
        \Else {\;break}
            \EndIf
        \EndWhile\;
    \Else 
        \State{\;\Return R}
        \EndIf
\EndWhile
\end{algorithmic}
\end{algorithm}

	
	











\section{Trace criterion as a feature selection criterion}\label{tracecri}

Trace criterion is a useful class separability measure for feature selection in classification tasks (more details in \cite{fukunaga2013introduction,johnson2002applied}). There are many equivalent versions. However, suppose that we have $C$ classes, and there are $n_i$ observation for the $i^{th}$ class, and let $\mathbf{x}_{ij}$ be the $j^{th}$ sample of the $i^{th}$ class.  Then, one way to define the criterion is
\begin{equation}
trace(S_w^{-1}S_b),
\end{equation}
where 
\begin{equation}
S_b = \sum_{i=1}^Cn_i(\bar{\boldsymbol{x} }_i-\bar{\boldsymbol{x} })(\bar{\boldsymbol{x} }_i-\bar{\boldsymbol{x} })'
\end{equation}
and 
\begin{equation}
S_w = \sum_{i=1}^C\sum_{j=1}^{n_i}( {\boldsymbol{x}  }_{ij}-\bar{\boldsymbol{x} }_i)({\boldsymbol{x} }_{ij}-\bar{\boldsymbol{x} }_i)'
\end{equation}
are the \textit{between-class scatter matrix} and \textit{within-class scatter matrix}, respectively. Here, $A'$ is the transpose of a matrix $A$, $\bar{\boldsymbol{x}}_i$ is the class mean of the $i^{{th}}$ class, and $\bar{\boldsymbol{x}}$ is the overall mean, i.e.,
\begin{align*}
    \bar{\boldsymbol{x}}_i &= \frac{1}{n_i}\sum_{j=1}^{n_i}(\bar{\boldsymbol{x}}_{ij}-\bar{\boldsymbol{x}}_i)\\
    \bar{\boldsymbol{x}} &= \frac{1}{\sum_{i=1}^Cn_i}\sum_{i=1}^C\sum_{j=1}^{n_i}(\bar{\boldsymbol{x}}_{ij}-\bar{\boldsymbol{x}}_i).
\end{align*}

This leads to the following result:
\begin{thm}
\begin{equation}\label{th1}
    trace(S_w^{-1}S_b) = \sum_{i=1}^C(\bar{\boldsymbol{x}}_i- \bar{\boldsymbol{x}})'S_w^{-1}(\bar{\boldsymbol{x}}_i- \bar{\boldsymbol{x}})
\end{equation}
\end{thm}
The proof of this statement is similar to the (population) mean and covariance matrix version in~\cite{johnson2002applied}. From this theorem, we can see that the trace criterion \ref{th1} can be considered as the sum of squared Mahalanobis distances from the class means to the overall means.
In addition, when the number of classes is two, the criterion \ref{th1} can be considered as the empirical estimation of KL-divergence of two multivariate normal distributions that have the same covariance matrix \cite{kong2014pairwise}. 

Since this criterion measures the separability of classes, we aim to maximize it. For notational simplicity, we write $\{R,f\}$ instead of $R\cup \{f\}$, and $t_{R, f}$ instead of  $t_{\{R, f\}}$. We then have the following theorem:

\begin{thm}\label{th2}
Let $R$ be the set of selected features, and $f$ be an arbitrary feature that does not belong to $R$.

Let $S_{Rf}$ be the value of $S_w$ when $\{R, f\}$ is the set of selected features, and $S_R$ be the value of $S_w$ when $R$ is the set of selected features. Partition
\begin{equation}
    S_{Rf} = \begin{pmatrix}
    S_R & v\\
    v'& u
    \end{pmatrix},
\end{equation}
where $ u \in \mathbb{R}^{+}, v \in \mathbb{R}^{|R| \times 1}, |R| $ is the cardinality of $R$.

If $u - v'S_R^{-1}v>0$ then
\begin{equation}
    t_{R,f}\ge t_R.
\end{equation}
Else, if $u - v'S_R^{-1}v<0$ then
\begin{equation}\label{eq7}
    t_{R,f}< t_R.
\end{equation}
\end{thm}
The proof of this theorem is given in Appendix \ref{pr2}. This theorem implies that the improvement or degradation of the model after adding a feature depends not only on the features themselves but also on the interaction between that feature and the features already in the model. In addition, from Equation \ref{eq7}, we can see that adding a feature may also reduce the class separability of a model. Hence, feature selection is needed to remove these features and the features that can only improve the model performance insignificantly. This property is desirable and not all feature selection criteria satisfy it. For example, for the mean square error (MSE), adding extra features, regardless of whether that feature is redundant or not, never increase MSE, as stated in the following theorem:

\begin{thm} \label{th3}
Let $MSE_X$ be the resulting mean squared error when we regress $Y$ based on $X$. Suppose $T$ contains arbitrary features to be added to the model and
\begin{equation}
    U = (X\;\;\;T).
\end{equation}
Let $MSE_U$ be the resulting mean squared error when we regress $Y$ based on $U$. If $M = (T'T - T'X(X'X)^{-1}X'T)^{-1}$ exists then,
$MSE_U \le MSE_X.$
\end{thm}
The proof of this theorem is given in Appendix \ref{pr3}.

\section{Parallel  Feature Selection  using  Trace  criterion (PFST Algorithm)}\label{methods}

\begin{algorithm}
\caption{\textbf{Parallel forward-backward algorithm with early dropping} }\label{alg:pfbaed}
\hspace*{\algorithmicindent} \textbf{Input:} Set A of all features, partitioned into feature blocks $F_1,..., F_b$, response $Y$, forward threshold $\alpha$, backward threshold $\beta $, early dropping threshold $\gamma$, maximum number of re-forward steps maxRef.\\
    \hspace*{\algorithmicindent} \textbf{Output:} List R of relevant features.
    
    \hspace*{\algorithmicindent}
    \textbf{Procedure:}
\begin{algorithmic}[1]
\State $f_b \gets \arg\max_f \{t_{\{f\}}: f\in F_b \}, b=1,...,B$ parallelly on B workers.
\State $R \gets \{f_1,...,f_B\}$.\\
 \textbf{\# Forward with forward-dropping stage:}
\While{$\bigcup_{1\le b\le B} {F}_b \neq \emptyset$}
    \State $f_b,F_b \gets OneForwardDropping(F_b,Y,\alpha,\gamma)$ (run parallel on $B$ workers).
    \State $R \gets R\cup \{f_1,..., f_B\}$ 
\EndWhile\\
 \textbf{\# Re-forward stage:}
\State Reset the selection pool to $A\setminus R$ and partition it into $F_1,..., F_B$.
 
	\State runs $\gets 0$
	\While{runs $<$ maxRef~\&~ $\bigcup_{1\le b\le B} {F}_b \neq \emptyset$}
	\State $f_b,F_b \gets$ OneReforward$(F_b,Y,\alpha)$ (run parallel on $B$ workers)
	\State $R \gets R\cup \{f_1,..., f_B\}$
	\State runs $\gets$ runs $+1$
	\EndWhile\\
\textbf{\# Backward stage:}
\State Partition $R$ into $B$ blocks of features.
\State $f_b \gets \arg \max_{f\in F_b} t_{R\setminus\{f\}}, b=1,...B$ parallelly on $B$ workers.
\State $f_r \gets \arg \max_{f\in \{f_1,...,f_B\}} t_{R\setminus \{f\}}$
\If{$t_R - t_{R\setminus\{f_r\} } <\beta$}
\State $R \gets R\setminus \{f_r\}$
\Else
\State break
\EndIf
\State \Return R
\end{algorithmic}
\end{algorithm}

This section details the PFST algorithm for parallel feature selection. In addition to the desirable properties of the trace criterion as discussed in the previous section, the method is based on the following observations: 

First, it is well known that forward feature selection is faster than backward feature selection because it sequentially adds the best relevant feature into the model. However, it is still computationally expensive. A potential way to speed up the process is to adapt the Early Dropping heuristic \cite{borboudakis2019forward}, in which features that are deemed unlikely to increase the performance of the model are discarded in subsequent iterations. In the original paper \cite{borboudakis2019forward}, the authors evaluate such possibilities by $p-$values. Yet, computing $p-$value using bootstrap is expensive. In our approach, we use the Early Dropping heuristic with the trace criterion as the evaluation tool instead. This does not require multiple iterations through the data for a forward step as using $p-$value for evaluation. 

Second, stepwise feature selection appears to be a remedy to the forward error in the forward selection. It adds features to the model sequentially as in the forward feature selection. In addition, after adding a new feature, this approach removes from the model feature(s) that is no longer important according to the feature selection criterion being used. However, there is a computational cost associated with the backward steps that remove unnecessary features. \cite{zhang2011adaptive} proposes an algorithm that takes a backward step only when the squared error is no more than half of the squared error decrease in the earlier forward steps. Yet, for a large-scale dataset, there is computational cost with checking a backward step, while there may not exist a redundant feature after a forward at all. In fact,  \cite{nguyen2019faster} provides a simulation example to show that a backward step is rarely taken in stepwise feature selection. 
Hence, in our approach, we add features until reaching a satisfactory model and then use backward feature selection to remove those redundant features.

Next, note that for single feature, $trace(S_w^{-1}S_b)$ reduces to 
\begin{equation}
t = \frac{\sum_{i=1}^C n_i(\bar{\boldsymbol{x} }_i-\bar{\boldsymbol{x} })^2}{\sum_{i=1}^C\sum_{j=1}^{n_i}(\boldsymbol{x} _{ij}-\bar{\boldsymbol{x} }_i)^2}
\end{equation}
and for normalized features, $S_b$ reduced to $S_b = \sum_{i=1}^C n_i \bar{x}_i \bar{x}_i'$. 
Let 
\begin{equation}
    t_R = trace(S^{-1}_wS_b)
\end{equation}
when only the set of features $R$ is included in the model.

For the sake of clarity, we added an overview of the descriptions for the used notations in Table \ref{tab_notations}. Finally, with all these heuristics and observations, the proposed PFST algorithm is presented in Algorithm 4. 

\begin{table}[htbp]
\caption{Table of notations}
\centering
    \begin{tabular}{|c|c|}
		\hline
		\textbf{Symbol} & \textbf{Description} \\\hline
		 $C$ & number of classes   \\

		$A'$& the transpose of matrix $A$ \\
		
		 $\bar{\boldsymbol{x}}_i$& the class mean of the $i^{th}$ class   \\
		
		 $\bar{\boldsymbol{x}}$& the overall mean   \\
		$R$ & the set of selected features \\
		 $\bar{\boldsymbol{x}}_{iRf}$ & the mean of the $i^{th}$ class when the selected features are $\{R,f\}$\\
		 $\bar{\boldsymbol{x}}_{Rf}$ &  the overall mean when the selected features are $\{R, f\}$\\
		
		 $n_i$ & number of observations in the $i^{th}$ class   \\
		
		 $x_{ij}$& the $j^{th}$ sample of the $i^{th}$ class   \\

		$S_{R}$ &The value of $S_w$ when $R$ is the set of selected features \\
		
		$t_{R}$ & $trace(S_w^{-1}S_b)$ when $R$ is the set of selected features \\
\hline
	\end{tabular}
	\label{tab_notations}
\end{table}

We firstly partition the features into $B$ blocks to assign to $B$ workers. At first, the set of selected features R is empty. Therefore, we compute $\arg\max_{f}\{t_f:f\in F_b\}, b = 1,...,B$ parallel on $B$ workers to produce an initial set $R$.

Next, we use the OneForwardDropping algorithm (Algorithm \ref{alg:ofd}) to sequentially add relevant features to the model and discard seemingly unimportant ones from subsequent iterations. To be more specific, the algorithm removes all the features in the pool that can not improve the trace criterion more than any amount of $\beta$. This helps reduce the computational cost of re-scanning through the features that temporarily do not seem to improve the model a lot compared to other features.
However, a major drawback of the forward with early dropping stage is that a seemingly useless feature at a step may be helpful later when combined with other features. Therefore, after these steps are finished, we run the re-forward stage, we reset the features pool to $P = A\setminus R$ and re-partition into B blocks, and conduct forward selection without early dropping. Since most of the important features are already in the model, this stage is not as computationally expensive as running forward selection parallelly when combined with the forward stage. We illustrate in the experiments that our method runs only in a fraction of time compared to the other techniques.

\begin{algorithm}
\caption{\textbf{OneForwardDropping} }\label{alg:ofd}
\hspace*{\algorithmicindent} \textbf{Input:} Feature block $F_b$, response $Y$, forward threshold $\alpha$, early dropping threshold $\gamma$.\\
    \hspace*{\algorithmicindent} \textbf{Output:} a relevant feature $f_b$ and the remaining features in the selection pool $F_B$ for worker B.
\begin{algorithmic}[1]
    \State $f_b \gets \arg\max_{f\in F_b}	t_{R,f}	$
	
	\If {$t_{R,f_b}-t_R<\alpha$}
	\State \# if there is no more relevant features, drop block by setting it to  $\emptyset$
	\State $F_b\gets \emptyset$
	\Else
	\State \# Early dropping:
	\State $D \gets \{f:t_{\{R,f\}}-t_R<\gamma  \}$
	\State $F_b \gets F_b\setminus \{D, f_b \} $
	\EndIf
\State \Return $f_b,F_b$
\end{algorithmic}
\end{algorithm}

	
	
	
		
	
	
		
	
	
	

Finally, note that the inclusion of new features may lead to some features that are included before becoming redundant, possibly due to feature interaction. Therefore, the last stage of PFST is the Backward Stage, which tries to remove the redundant features that forward-dropping and re-forward stages may have made before. In this stage, we conduct backward feature selection steps to remove the redundant features that remained in the model. Again, this is to avoid the computational cost of checking for a backward move after every inclusion of a new feature. This is different from how backward steps are conducted in stepwise selection.

\begin{algorithm}
\caption{\textbf{OneReforward} }\label{alg:or}
\hspace*{\algorithmicindent} \textbf{Input:} Feature block $F_b$, response $Y$, forward threshold $\alpha$.\\
    \hspace*{\algorithmicindent} \textbf{Output:} A relevant feature $f_b$ and the remaining features in the selection pool $F_B$ for worker B.
\begin{algorithmic}[1]
    \State $f_b \gets \arg\max_{f\in F_b}	t_{R,f}	$
    \If{$t_{R,f_b}-t_R<\alpha$}
    \State \# if there is no more relevant features, drop block by setting it to  $\emptyset$:
	
	\State $F_b\gets \emptyset$
    \Else
    \State $F_b \leftarrow F_b\setminus\{f_b\}$
    \EndIf
\State \Return $f_b,F_b$
\end{algorithmic}
\end{algorithm}

	
	
	
	
	
	
	
	

Note that a higher $\gamma$ will result in more feature dropping and less re-scanning during forward-dropping moves. However, depending on the data and the chosen criterion, we may prefer to use a lower $\gamma$ for high dimensional data. The reason is that a higher  $\gamma$ may result in too many feature droppings, which implies that much fewer features have chances to get into the model during the forward-dropping moves. This causes the forward-dropping moves to terminate early, and we have to re-scan a large number of features during the re-forward stage.

It is also important to point out that after the forward-dropping steps are the re-forward steps. Therefore, after the forward-dropping steps, the algorithm ranks the importance of the features. It is possible to specify the maximum number of features to be included in the model if one wishes for a smaller set of features than what the thresholds may produce. Hence, we can specify the maximum number of features to be included in the model for the re-forward stage.  

Last but not least, if the number of features selected after the re-forward stage is not significantly large, one should use the regular backward feature selection instead of parallel backward feature selection. This is because when the number of features in the model is small, parallelism makes the process run slower. 



 


\section{Experiments}\label{exper}
\begin{table}[htbp]
\caption{The description of datasets that are used in our experiments}
\begin{center}
    \begin{tabular}{|c|c|c|c|}
		\hline
		Dataset & \# Classes & \# Features & \# Samples \\
		\hline
		Breast cancer & $2$ & $30$ & $569$ \\
		\hline
		Parkinson & $2$ & $754$ & $756$ \\
		\hline
		Mutants & $2$  & $5408$ & $31419$\\
		\hline
		Gene & $5$ & $20531$ & $801$ \\
		\hline
		Micromass & $10$ & $1087$ & $360$\\
		\hline
	\end{tabular}
	\label{table_info_datasets}
\end{center}
\end{table}

To illustrate the performance of our approach, we compare our method with Parallel Sequential Forward Selection (PSFS) \cite{scikit-learn}, Parallel Sequential Backward Selection (PSBS) \cite{scikit-learn}, Parallel Support Vector Machine Feature Selection based on Recursive Feature Elimination with Cross-Validation (PSVMR) \cite{guyon2002gene}, and Parallel Mutual Information-based Feature Selection (PMI) \cite{bennasar2015feature}. 

\subsection{Datasets \& Settings}

The experiments are done on the datasets from the Scikit-learn library \cite{scikit-learn} and UCI Machine Learning repository \cite{Dua:2019}. The details of these datasets are given in Table \ref{table_info_datasets}.  

\begin{table*}[htbp]
\caption{5-fold misclassification rate}
\resizebox{\textwidth}{!}{%
\begin{tabular}{|l|c|c|c|c|c|c|c|}
\hline
\textbf{Datasets} &
\textbf{\# Selected Features} & 
  {\textbf{PFST (our)}} &
  {\textbf{PSFS}} &
  {\textbf{PSBS}} &
  {\textbf{PSVMR}} &
  {\textbf{PMI}} &
  \textbf{Full Features} \\ \hline
\textbf{Breast cancer} &
  $3$ &
  {$\boldsymbol{0.042}$} &
  {$0.111$} &
  {$0.074$} &
  {$0.051$} &
  {$0.076$} &
  $0.042$ \\ \hline
\textbf{Parkinson} &
  $11$ &
  {$\boldsymbol{0.112}$} &
  {$0.234$} &
  {NA} &
  {NA} &
  {$0.181$} &
  $0.362$ \\ \hline
\textbf{Mutants} &
  $6$ &
  {\textbf{0.008}} &
  {NA} &
  {NA} &
  {NA} &
  {NA} &
  0.010 \\ \hline
\textbf{Gene} &
  $12$ &
  {$\boldsymbol{0.006}$} &
  {$0.009$} &
  {NA} &
  {NA} &
  {$0.007$} &
  $0.042$ \\ \hline
\textbf{Micromass} &
  $19$ &
  {$0.115$} &
  {$0.310$} &
     {$0.218$} &
  {$\boldsymbol{0.096}$} &
  {$0.228$} &
  $0.129$ \\ \hline
\end{tabular}
\label{table_error}
}
\end{table*}

\begin{table*}[htbp]
\caption{Running time and number of selected features}
\resizebox{\textwidth}{!}{%
\begin{tabular}{|l|c|c|ccccc|}
\hline
\multicolumn{1}{|c|}{\multirow{2}{*}{\textbf{Datasets}}} &
  \multirow{2}{*}{\textbf{\begin{tabular}[c]{@{}c@{}}\# selected\\ features\end{tabular}}} & \multicolumn{1}{c|}{\multirow{2}{*}{\textbf{\# Features}}} &
  \multicolumn{5}{c|}{\textbf{Running Time (s)}}  \\ \cline{4-8} 
\multicolumn{1}{|c|}{} &
   & &
  \multicolumn{1}{c|}{\textbf{PFST (our)}} &
  \multicolumn{1}{c|}{\textbf{PSFS}} &
  \multicolumn{1}{c|}{\textbf{PSBS}} &
  \multicolumn{1}{c|}{\textbf{PSVMR}} &
  \multicolumn{1}{c|}{\textbf{PMI}} 
 \\ \hline
\textbf{Breast cancer} &
  3 & 30& 
  \multicolumn{1}{c|}{\textbf{0.095}} &
  \multicolumn{1}{c|}{1.403} &
  \multicolumn{1}{c|}{8.268} &
  \multicolumn{1}{c|}{12.470} &
  \multicolumn{1}{c|}{0.646}    \\ \hline
\textbf{Parkinson} &
  11 & 754&
  \multicolumn{1}{c|}{\textbf{3.219}} &
  \multicolumn{1}{c|}{163.32} &
  \multicolumn{1}{c|}{NA} &
  \multicolumn{1}{c|}{NA} &
  \multicolumn{1}{c|}{77.269}   \\ \hline
\textbf{Mutants} &
  6 & 5408 &
  \multicolumn{1}{c|}{\textbf{674.702}} &
  \multicolumn{1}{c|}{NA} &
  \multicolumn{1}{c|}{NA} &
  \multicolumn{1}{c|}{NA} &
  \multicolumn{1}{c|}{NA} \\ \hline
\textbf{Gene} &
  12 & 20531 & 
  \multicolumn{1}{c|}{\textbf{172.386}} &
  \multicolumn{1}{c|}{5350.14} &
  \multicolumn{1}{c|}{NA} &
  \multicolumn{1}{c|}{NA} &
  \multicolumn{1}{c|}{2706.45}   \\ \hline
\textbf{Micromass} &
  19 & 1087&
  \multicolumn{1}{c|}{\textbf{16.1}} &
  \multicolumn{1}{c|}{252.9} &
  \multicolumn{1}{c|}{10561.3} &
  \multicolumn{1}{c|}{46.909} &
  \multicolumn{1}{c|}{144.684} 
   \\ \hline
\end{tabular}
\label{tab_time}
}
\end{table*}

For the p53 mutants dataset, we eliminated a row of all null values. In addition, we added a small amount of noise to the Gene expression cancer RNA-Seq dataset and the p53 mutants dataset to avoid inversion error resulting from matrix singularity when running PSFT.

For PSFT, the thresholds used are $\alpha= \gamma = 0.05, \beta=0.01$. For PSBS and PSFS, we used the K-nearest neighbors algorithm with $K=3$ as the estimator. For PSVMR, we used linear kernel for PSVMR and \lq\lq JMI\rq\rq~for PMI. For a fair comparison, we force other feature selection techniques to select the same number of features as PSFT.

We run the experiments on an AMD Ryzen 7 3700X CPU with 8 Cores and 16 processing threads, 3.6GHz and 16GB RAM. After selecting the relevant features, we classify the samples using \textit{linear discriminant analysis (LDA)} and report the 5-fold misclassification rate in Table \ref{table_error}. In addition, we present the running time and the number of selected features compared to the total number of features in Table \ref{tab_time}.

We terminate an experiment if no result is produced after five hours of running or when having the out-of-memory issue, and denote this as NA in the result Tables \ref{table_error} and \ref{tab_time}).

\subsection{Results and discussion}

From the Tables \ref{table_error} and \ref{tab_time}, we can see that our PFST feature selection method has excellent performance not only in terms of speed but also the accuracy after doing classification. For example, in the breast cancer dataset, PFST was able to produce a set of 3 relevant features out of 30 features in only 0.095 seconds (s), while PSBS and PSVMR need 8.268s and 12.470s, respectively. Even though being the fastest feature selection method, it can achieve the best classification result with a misclassification rate of only 0.042, which is as good as classification using all available features. Moreover, we can observe that the performance of PFST is better than PSBS, which is the parallel version of backward feature selection.

Interestingly, for the Parkinson dataset,  PFST selects only 11 out of 754 features. Still, the misclassification rate is the lowest (0.112), which is only  33\% of the misclassification rate of classification using all available features (0.362), and 66.87\% of the misclassification rate of the next best performer PMI (0.181). This implies that removing redundant features may boost the performance of the classifier significantly. For the running time, note that PFST takes only 3.219s to get 11 features from 754 features, while PMI takes 77.269s, PSFS takes 163.32s. PSBS and PSVMR can not get the results when running for more than 5 hours.

In the dataset with the most features, the Gene Expression Cancer RNA-Seq dataset, PFST reduced 20531 features to 12 features within 3 minutes and got the best misclassification rate of only 0.006. Although PSFS and PMI have also gotten good performance with a misclassification rate of only 0.009 and 0.007, respectively, they need much longer than PFST, while PSBS and PSVMR cannot get the result within 5 hours.

Even though PFST is the fastest algorithm among the ones used in the experiments, it outperforms other approaches in four out of five datasets in terms of classification results.{ For the Micromass dataset, the best performer for the classification is PSVMR, and the next best is PFST. However, PSVMR is only slightly better than PFST in terms of classification result (1.9\% lower misclassification rate), but its running time is almost three times longer than that of PFST (46.909s and 16.1s, respectively).}

\section{Conclusion} \label{conclu}
This paper presents a novel parallel feature selection approach for classification, called PFST, for feature selection on large scale datasets. The method uses the trace criterion, i.e., a criterion with many desirable properties, to evaluate feature usefulness. We evaluate the approach via various experiments and datasets using Linear Discriminant Analysis as the classifier on selected features. The experiments show that our PFST method can select relevant features in a fraction amount of time compared to other compared state of the art approaches. In addition, the classifier trained on the features selected by PFST achieves better accuracy than those selected by other methods in four out of five cases and better than the model that is fitted on all available features. However, the trace criterion can only be used for continuous features. Therefore, in the future, it would be desirable to explore how to extend this work to the case where there are categorical features in the model as well. 

\bibliographystyle{IEEEtran}
\bibliography{bbli}
\appendix
\subsection{Proof of Theorem \ref{th2}}\label{pr2}
Let $\bar{\boldsymbol{x}}_{iR}, \bar{\boldsymbol{x}}_R$ be the mean of the $i^{th}$ class, and the overall mean when a set $R$ of features are included in the model, respectively. 
Similarly, let $\bar{\boldsymbol{x}}_{iRf}, \bar{\boldsymbol{x}}_{Rf}$ be the mean of the $i^{th}$ class, and the overall mean when a set $\{R, f\}$ of features are included in the model, respectively.  Next, note that 
\begin{equation}
    \bar{\boldsymbol{x}}_{iRf} = \begin{pmatrix}
    \bar{\boldsymbol{x}}_{iR}\\\bar{\boldsymbol{x}}_{if}
    \end{pmatrix},
\end{equation}
and 
\begin{equation}
    \bar{\boldsymbol{x}}_{Rf} = \begin{pmatrix}
    \bar{\boldsymbol{x}}_{R}\\\bar{\boldsymbol{x}}_{f}
    \end{pmatrix},
\end{equation}
where $\bar{\boldsymbol{x}}_{if}, \bar{\boldsymbol{x}}_f$ is the $i^{th}$ class mean, and the overall mean of the feature $f$, respectively.

If $u\neq v'S_R^{-1}v$ then $M = (u-v'S_R^{-1}v)^{-1}$ exists. Hence, 
\begin{equation}
    S_{Rf}^{-1} = \begin{pmatrix}
    S_R^{-1}+S_R^{-1}vMv'S_R^{-1} & -S_R^{-1}vM\\
    -Mv'S_R^{-1}&M
    \end{pmatrix}
\end{equation}

Next, let
\begin{align*}
    \eta_{iRf} &= (\bar{\boldsymbol{x}} _{iRf}-\bar{\boldsymbol{x}}_{Rf})'S_{Rf}^{-1}(\bar{\boldsymbol{x}} _{iRf}-\bar{\boldsymbol{x}}_{Rf})\\
    &= \eta_{iR}+(\bar{\boldsymbol{x}}_{iR}-\bar{\boldsymbol{x}}_R)'S_R^{-1}vMv'S_R^{-1}(\bar{\boldsymbol{x}}_{iR}-\bar{\boldsymbol{x}}_R)\\
    & - 2(\bar{\boldsymbol{x}}_{iR}-\bar{\boldsymbol{x}}_R)'S_R^{-1}vM(\bar{\boldsymbol{x}}_{if}-\bar{\boldsymbol{x}}_f)\\
    &+ (\bar{\boldsymbol{x}}_{if}-\bar{\boldsymbol{x}}_f)'M(\bar{\boldsymbol{x}}_{if}-\bar{\boldsymbol{x}}_f),
\end{align*}
where 
$$\eta_{iR} = (\bar{\boldsymbol{x}} _{iR}-\bar{\boldsymbol{x}}_{R})'S_{R}^{-1}(\bar{\boldsymbol{x}} _{iR}-\bar{\boldsymbol{x}}_{R}).$$
Note that $ M\in \mathbb{R}, M\neq 0$. Hence, 
\begin{align*}
    \frac{\eta_{iRf}-\eta_{iR}}{M}     &=
    (\bar{\boldsymbol{x}}_{iR}-\bar{\boldsymbol{x}}_R)'S_R^{-1}vv'S_R^{-1}(\bar{\boldsymbol{x}}_{iR}-\bar{\boldsymbol{x}}_R)\\
    & - 2(\bar{\boldsymbol{x}}_{iR}-\bar{\boldsymbol{x}}_R)'S_R^{-1}v(\bar{\boldsymbol{x}}_{if}-\bar{\boldsymbol{x}}_f)\\
    &+ (\bar{\boldsymbol{x}}_{if}-\bar{\boldsymbol{x}}_f)'(\bar{\boldsymbol{x}}_{if}-\bar{\boldsymbol{x}}_f).
\end{align*}
Applying Cauchy-Schwarz inequality $||a||^2+||b||^2\ge 2 \langle a,b\rangle$, with  $a = (\bar{\boldsymbol{x}}_{iR}-\bar{\boldsymbol{x}}_R)S_R^{-1}v$, and $b = (\bar{\boldsymbol{x}}_{if}-\bar{\boldsymbol{x}}_f)$, we see that
\begin{equation*}
   \frac{\eta_{iRf}-\eta_{iR}}{M} \ge 0.
\end{equation*}
Since, 
\begin{equation*}
    t_{R,f} - t_{R} = \sum_{i=1}^C (\eta_{iRf}- \eta_{iR}),  
\end{equation*}
it follows that if $u - v'S_R^{-1}v>0$ then
\begin{equation*}
    t_{R,f}\ge t_R,
\end{equation*}
and if $u - v'S_R^{-1}v<0$ then
\begin{equation*}
    t_{R,f}< t_R.
\end{equation*}

\subsection{Proof of Theorem \ref{th3}}\label{pr3}
Let $H = X(X'X)^{-1}X'$ then it is known from ~\cite{kutner2005applied} that $I-H$ is idempotent, i.e.,
\begin{equation}
    (I-H)^2 = I-H
\end{equation}
 Hence, the sum of square when regressing $Y$ over $X$ is
\begin{align*}
    SSE_X &= (Y-\hat{Y}_X)'(Y-\hat{Y}_X)\\
    &= Y'(I-H)Y
\end{align*}
where $\hat{Y}_X$ is the predicted outcome when regressing $Y$ over $X$.

Note that M is symmetric, and let $\hat{Y}_U$ is the predicted outcome when regressing $Y$ over $U$. Hence,
\begin{align*}
\small
    &\;\;\;\;\;(U'U)^{-1}U'Y \\
    &= HY+HTMT'HY-HTMT'Y - TMT'HY+TMT'Y\\
     &= HY-(I-H)TMT'+(I-H)TMT'Y.
\end{align*}
This implies
\begin{align*}
    &\;\;\;\;\;Y-\hat{Y}_U \\
    &= Y-\hat{Y}_X+(I-H)TMT'HY-(I-H)TMT'Y\\
    &= Y- \hat{Y}_X-(I-H)TMT'(I-H)Y
\end{align*}
 Hence, the sum of square error when regressing $Y$ over $U$ is
\begin{align*}
    SSE_U &= (Y-\hat{Y}_U)'(Y-\hat{Y}_U)\\
    &= SSE_X-2Y'(I-H)TMT'(I-H)(Y-\hat{Y}_X)\\
    &\;\;\;\;+Y'(I-H)TMT'(I-H)TMT'(I-H)Y
\end{align*}
In addition, note that $(I-H)TMT'$ is idempotent, and that $Y'(I-H)TMT'(I-H)Y\ge 0$. Therefore, 
\begin{align*}
    &\;\;\;\;\;Y'(I-H)TMT'(I-H)TMT'(I-H)Y\\
    &=Y'(I-H)TMT'(I-H)Y\\
    &\le 2Y'(I-H)TMT'(I-H)Y.
\end{align*}
Hence
\begin{equation*}
    SSE_U\le SSE_X
\end{equation*}
and the proof follows.


\end{document}